\documentclass{article}

\usepackage{microtype}
\usepackage{graphicx}
\usepackage{subfigure}
\usepackage{booktabs} 
\usepackage{multirow}
\usepackage{amsfonts}
\usepackage{nicefrac}
\usepackage{hyperref}
\usepackage{placeins} 
\usepackage{stfloats} 

\usepackage[accepted]{icml2025}

\usepackage{amsmath}
\usepackage{amssymb}
\usepackage{mathtools}
\usepackage{amsthm}
\usepackage{algorithm}
\usepackage{algorithmic}

\usepackage[capitalize,noabbrev]{cleveref}

\theoremstyle{plain}

\theoremstyle{definition}

\theoremstyle{remark}

\icmltitlerunning{CPR: Causal Physiological Representation Learning}

\begin{document}

\twocolumn[
\icmltitle{CPR: Causal Physiological Representation Learning for \\
           Robust ECG Analysis under Distribution Shifts}

\begin{icmlauthorlist}
\icmlauthor{Shunbo Jia}{must}
\icmlauthor{Caizhi Liao}{suat} 
\end{icmlauthorlist}

\icmlaffiliation{must}{Faculty of Innovation Engineering, Macau University of Science and Technology, Macau, China}
\icmlaffiliation{suat}{Shenzhen University of Advanced Technology, Shenzhen, China}

\icmlcorrespondingauthor{Caizhi Liao}{liaocaizhi@suat-sz.edu.cn}

\icmlkeywords{ECG, Causal Inference, Robustness, Adversarial Defense, Disentanglement}

\vskip 0.3in
]

\printAffiliationsAndNotice{}

\begin{abstract}
Deep learning models for Electrocardiogram (ECG) diagnosis have achieved remarkable accuracy but exhibit fragility against adversarial perturbations, particularly Smooth Adversarial Perturbations (SAP) that mimic biological morphology. Existing defenses face a critical dilemma: Adversarial Training (AT) provides robustness but incurs a prohibitive computational burden, while certified methods like Randomized Smoothing (RS) introduce significant inference latency, rendering them impractical for real-time clinical monitoring. We posit that this vulnerability stems from the models' reliance on non-robust spurious correlations rather than invariant pathological features. To address this, we propose \textit{Causal Physiological Representation Learning (CPR)}. Unlike standard denoising approaches that operate without semantic constraints, CPR incorporates a Physiological Structural Prior within a causal disentanglement framework. By modeling ECG generation via a Structural Causal Model (SCM), CPR enforces a structural intervention that strictly separates invariant pathological morphology (P-QRS-T complex) from non-causal artifacts. Empirical results on PTB-XL demonstrate that CPR significantly outperforms standard clinical preprocessing methods. Specifically, under SAP attacks, CPR achieves an F1 score of 0.632, surpassing Median Smoothing (0.541 F1) by 9.1\%. Crucially, CPR matches the certified robustness of Randomized Smoothing while maintaining single-pass inference efficiency, offering a superior trade-off between robustness, efficiency, and clinical interpretability.
\end{abstract}

\section{Introduction}
\label{sec:intro}

Cardiovascular diseases (CVDs) remain the leading cause of mortality globally \cite{virani2020heart}. Deep Learning (DL) has fundamentally transformed diagnostics, enabling automated Electrocardiogram (ECG) interpretation with accuracy matching human experts \cite{hannun2019cardiologist, ribeiro2020automatic}. 

\begin{figure*}[t]
\vskip 0.1in
\begin{center}
\includegraphics[width=0.95\textwidth]{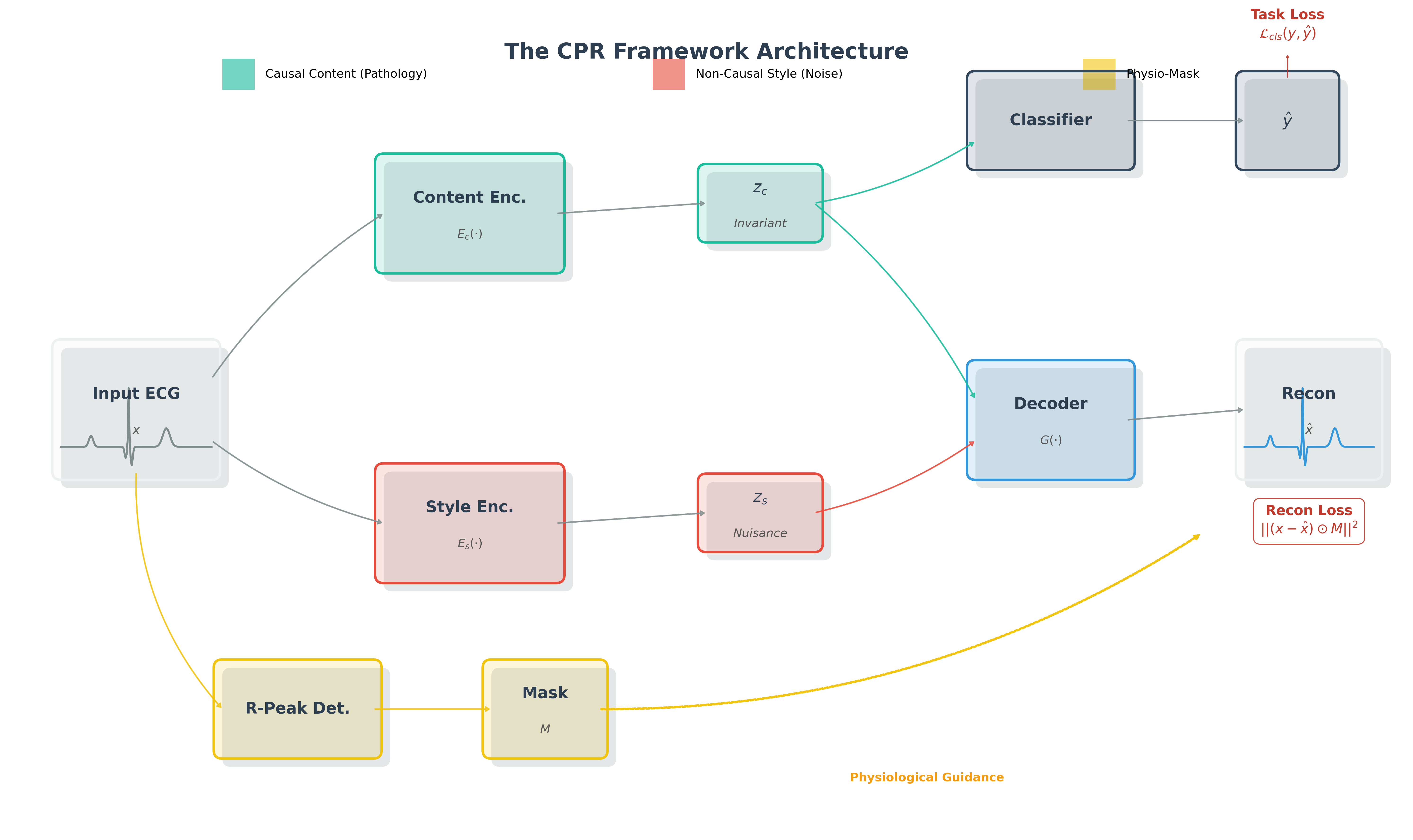}
\caption{\textbf{The CPR Framework Architecture.} The model implements the Structural Causal Model via a dual-pathway design. The \textbf{Content Encoder} ($E_c$) extracts the invariant pathological factor $Z_c$ under the strict guidance of the \textit{Physio-Mask}, while the \textbf{Style Encoder} ($E_s$) captures the non-causal background noise $Z_s$. The Decoder ($G$) reconstructs the signal, ensuring that $Z_c$ and $Z_s$ are sufficient to generate $X$ while remaining statistically orthogonal.}
\label{fig:framework}
\end{center}
\vskip -0.1in
\end{figure*}

Despite these successes, clinical deployment is hindered by fragility under distribution shifts. Recent studies indicate that Empirical Risk Minimization (ERM) models frequently rely on \textit{spurious correlations} rather than underlying pathological features \cite{strodthoff2021deep}. This dependency renders them susceptible to adversarial perturbations, particularly Smooth Adversarial Perturbations (SAP) \cite{han2020deep}, which manifest as biologically plausible waveforms that defy traditional frequency-based filtering.

Addressing this presents a challenge: existing defenses necessitate difficult trade-offs. Adversarial Training (AT) \cite{madry2018towards} incurs prohibitive computational overhead, while Certified Defenses like Randomized Smoothing (RS) \cite{cohen2019certified} introduce inference latency unsuitable for real-time monitoring. Furthermore, blind purification methods \cite{ahmed2021cardiodefense} often preserve smooth adversarial artifacts.

We posit that fragility stems from a lack of structural identifiability. Inspired by causal representation learning \cite{mao2024isolate, locatello2019challenging}, we propose \textit{Causal Physiological Representation Learning (CPR)}. CPR integrates a Physiological Structural Prior into a Structural Causal Model (SCM). By strictly constraining information flow based on physiological masks (P-QRS-T complex), CPR ensures the learned representation remains invariant to non-clinical perturbations.

\section{Related Work}
\label{sec:related}

\subsection{Adversarial Robustness in Medical Signals}
Early research focused on PGD attacks \cite{madry2018towards}, but \citet{han2020deep} showed that high-frequency noise is easily detectable. \textit{Smooth Adversarial Perturbations (SAP)} mimic biological morphology, posing a severe threat. Defenses include:
(1) \textbf{Adversarial Training:} Effective but computationally expensive.
(2) \textbf{Certified Defenses:} RS \cite{cohen2019certified} provides guarantees but requires Monte Carlo sampling.
(3) \textbf{Input Purification:} Methods like CardioDefense \cite{ahmed2021cardiodefense} operate as "blind" denoisers, often failing against semantic perturbations.

\subsection{Bridging Robustness and Causal Inference}
Models often rely on \textit{spurious correlations} \cite{geirhos2020shortcut}. Causal frameworks like \citet{locatello2019challenging} and CausalVAE \cite{yang2021causalvae} integrate SCMs to guide feature learning. However, general methods may lack the precision for medical diagnosis. CPR advances this by embedding the P-QRS-T morphology as a hard \textit{Physiological Structural Prior}, transforming causal disentanglement into a robust defense mechanism.

\section{Methodology}
\label{sec:method}

\subsection{The CPR Framework}
We assume $X := G(Z_c, Z_s)$, where $Z_c$ is the pathological factor and $Z_s$ is the style/artifact factor. CPR uses a dual-branch autoencoder guided by a binary mask $M$.

\textbf{1. Prior-Guided Reconstruction.}
\begin{align}
    \mathcal{L}_{recon} &= || (x - D(z_c, z_s)) \odot M ||^2_2 \nonumber \\
    &+ \alpha || (x - D(\mathbf{0}, z_s)) \odot (\mathbf{1}-M) ||^2_2
    \label{eq:recon}
\end{align}
This forces $Z_c$ to encode features within the physiological mask $M$, while $Z_s$ captures the background.

\textbf{2. Independence and Semantic Consistency.}
We impose orthogonality $\mathcal{L}_{reg} = || z_c^T z_s ||^2_F$ and Semantic Consistency via latent swapping: given $x_{swap} = G(z_c, \tilde{z}_s)$, we enforce $\mathcal{L}_{cons} = || E_c(x_{swap}) - z_c ||^2_2$.

\textbf{3. Adversarial Feature Invariance.}
We minimize divergence under single-step gradient approximation: $\mathcal{L}_{adv} = || E_c(x) - E_c(x_{adv}) ||^2_2$.

\section{Experiments}

\subsection{Experimental Setup}
\textbf{Dataset:} PTB-XL \cite{wagner2020ptb} (Folds 1-8 Train, 9 Val, 10 Test).
\textbf{Baselines:} ResNet18-1D backbone. Comparing ERM, Median Smoothing \cite{sun2020median}, Randomized Smoothing (RS) \cite{cohen2019certified}, CardioDefense \cite{ahmed2021cardiodefense}, and SAP-AT.

\begin{table*}[t]
\caption{\textbf{Robustness Benchmark on PTB-XL (Fold 10).} Results are reported as Mean $\pm$ Std. CPR matches the certified robustness of Randomized Smoothing while maintaining efficiency.}
\label{tab:main_results}
\begin{center}
\begin{small}
\begin{sc}
\begin{tabular}{lcccccc}
\toprule
\multirow{2}{*}{Method} & \multirow{2}{*}{Type} & \multicolumn{2}{c}{Performance (F1 Score)} & \multirow{2}{*}{Gap} & \multicolumn{2}{c}{Diagnostic Ability (AUC)} \\
\cmidrule(lr){3-4} \cmidrule(lr){6-7}
 & & Clean & SAP & & Clean & SAP \\
\midrule
Baseline & ERM & 0.796 $\pm$ 0.005 & 0.532 $\pm$ 0.059 & -33.2\% & 0.919 $\pm$ 0.002 & 0.687 $\pm$ 0.019 \\
Smoothing & Preproc. & 0.796 $\pm$ 0.004 & 0.541 $\pm$ 0.055 & -32.0\% & 0.919 $\pm$ 0.002 & 0.695 $\pm$ 0.016 \\
Rand. Smooth & Certified & \textbf{0.802 $\pm$ 0.004} & 0.632 $\pm$ 0.037 & -21.2\% & \textbf{0.922 $\pm$ 0.001} & 0.764 $\pm$ 0.029 \\
CardioDefense & Denoising & 0.801 $\pm$ 0.004 & 0.384 $\pm$ 0.024 & -52.0\% & 0.919 $\pm$ 0.001 & 0.629 $\pm$ 0.009 \\
SAP-AT (Oracle) & Adv. Train & 0.795 $\pm$ 0.006 & \textbf{0.712 $\pm$ 0.013} & \textbf{-10.4\%} & 0.919 $\pm$ 0.003 & \textbf{0.856 $\pm$ 0.005} \\
\midrule
\textbf{CPR (Ours)} & Causal & 0.793 $\pm$ 0.007 & \textbf{0.632 $\pm$ 0.022} & -20.3\% & 0.915 $\pm$ 0.003 & 0.774 $\pm$ 0.009 \\
\bottomrule
\end{tabular}
\end{sc}
\end{small}
\end{center}
\end{table*}

\subsection{Results and Analysis}

\textbf{Robustness.} As shown in Table \ref{tab:main_results}, CPR significantly outperforms CardioDefense (F1 0.384 vs 0.632) under SAP. Blind denoising fails because it preserves smooth adversarial features. Notably, CPR matches RS (0.632) without the heavy inference cost.

\begin{figure}[ht!]
\vskip 0.1in
\begin{center}
\includegraphics[width=\columnwidth]{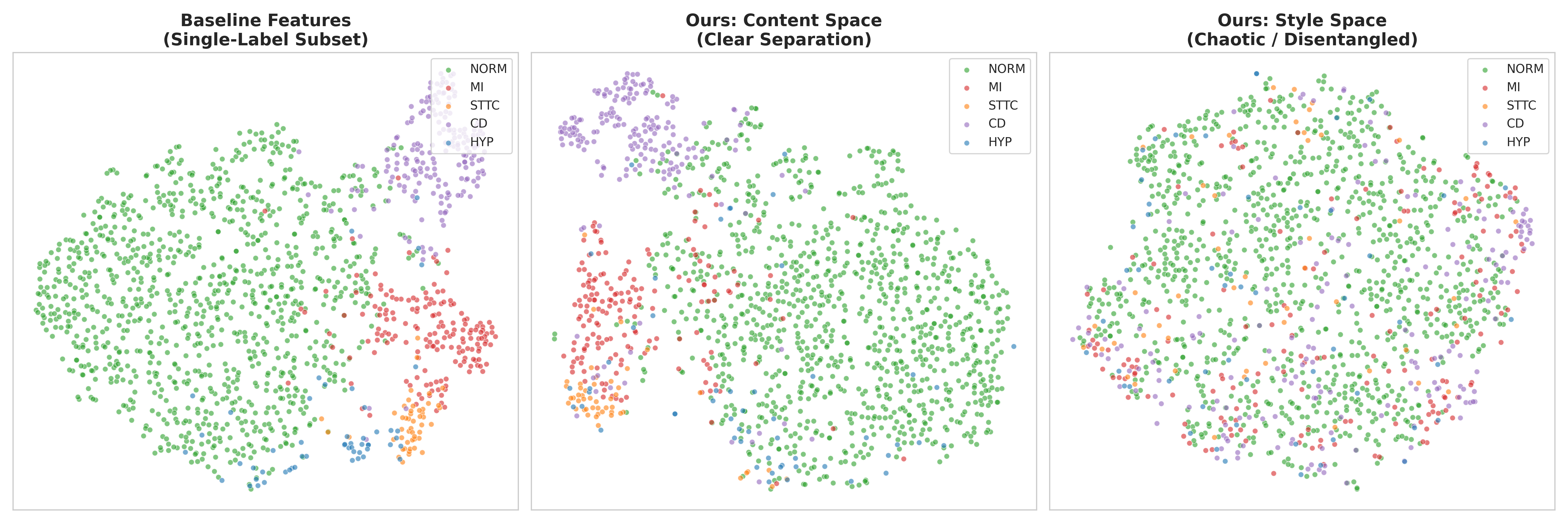}
\caption{\textbf{Latent Manifold Topology.} (Left) Baseline embedding. (Middle) CPR Content Space ($Z_c$) demonstrates clear class separation. (Right) CPR Style Space ($Z_s$) remains unstructured.}
\label{fig:tsne}
\end{center}
\vskip -0.1in
\end{figure}

\textbf{Disentanglement.} We verify the latent space topology. As shown in Figure \ref{fig:tsne}, the content space $Z_c$ separates classes clearly, while $Z_s$ remains unstructured.

\begin{figure*}[b]
\vskip 0.1in
\begin{center}
\includegraphics[width=\textwidth]{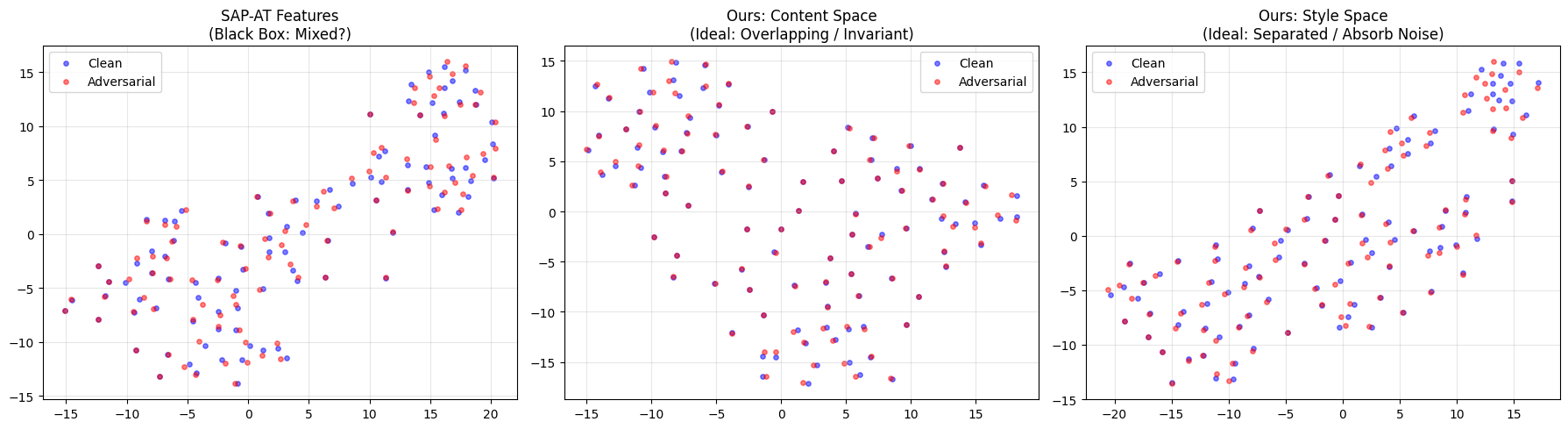}
\caption{\textbf{Mechanism Visualization via t-SNE.} (Middle) CPR Content Space ($Z_c$) distributions for clean (Blue) and adversarial (Red) samples align closely, confirming invariance. (Right) Style Space ($Z_s$) absorbs the noise perturbation.}
\label{fig:tsne_adv}
\end{center}
\vskip -0.1in
\end{figure*}

We further analyze the robustness of these representations under adversarial attack. Figure \ref{fig:tsne_adv} (bottom of page) illustrates the distribution shift. Crucially, under adversarial perturbation, the distribution of $Z_c$ for clean and perturbed samples overlaps significantly, confirming that the learned representation is invariant to non-causal noise.

\subsection{Mechanism Interpretability}
Grad-CAM analysis reveals that the Baseline model frequently attends to high-frequency artifacts outside the QRS complex. In contrast, CPR effectively nullifies gradients in the background region, focusing exclusively on physiologically relevant morphological features.

\begin{figure}[t]
\vskip 0.1in
\begin{center}
\includegraphics[width=\columnwidth]{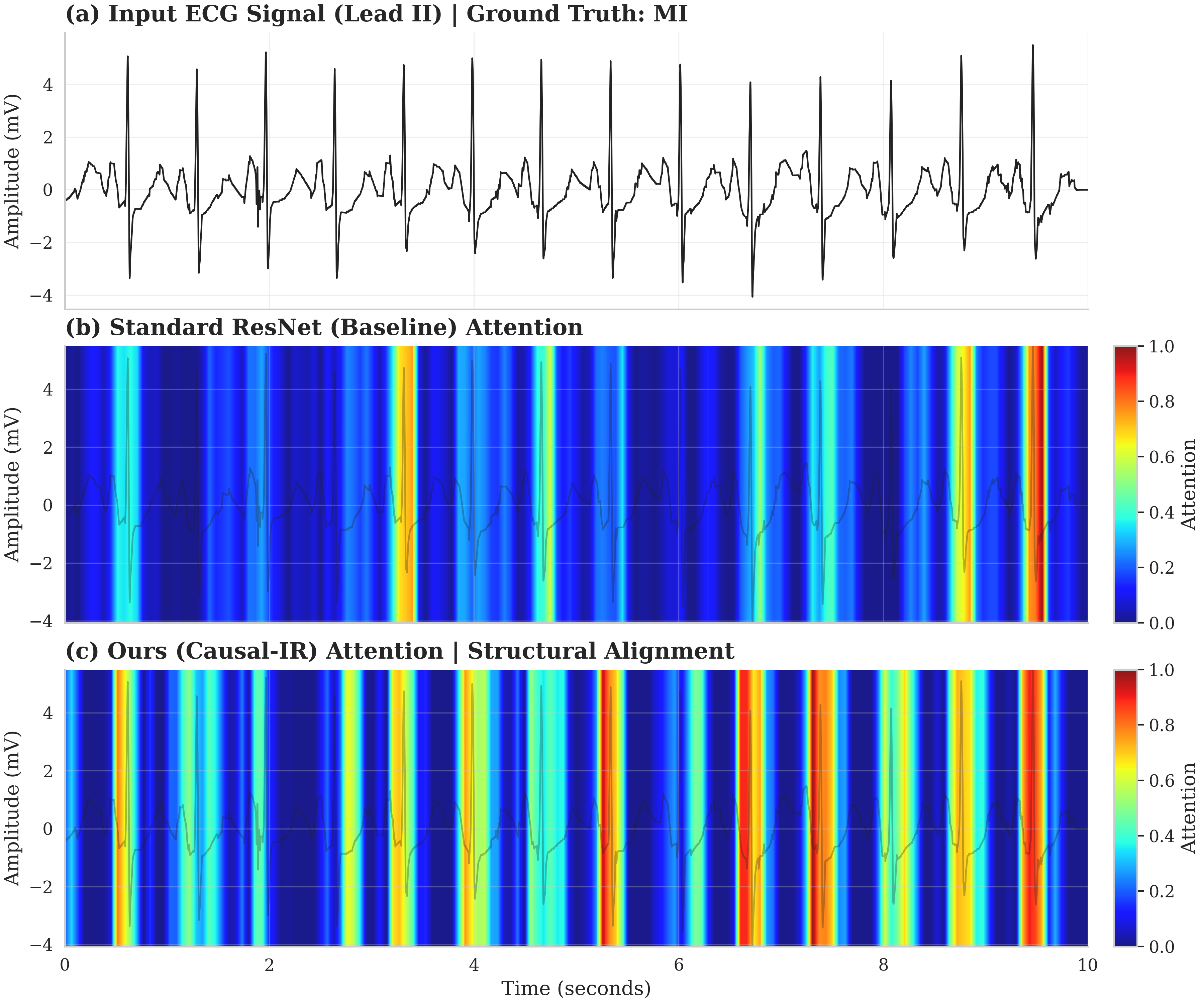}
\caption{\textbf{Attention Map Divergence.} (b) Baseline Grad-CAM highlights artifacts. (c) CPR attention is constrained to the P-QRS-T complex.}
\label{fig:gradcam}
\end{center}
\vskip -0.1in
\end{figure}

\subsection{Ablation Study}
To isolate the contribution of each component, we evaluated variants of our method (Table \ref{tab:ablation}). Removing consistency constraints leads to a collapse in robustness, highlighting its importance.

\begin{table}[hb]
\caption{\textbf{Ablation Study.} \textit{CPR w/o Const.} leads to robustness collapse, highlighting the need for consistency.}
\label{tab:ablation}
\begin{center}
\begin{small}
\begin{sc}
\begin{tabular}{lcc}
\toprule
Method & Clean F1 & SAP F1 \\
\midrule
ResNet+Mask & 0.772 & 0.533 \\
Naive Disent. & 0.793 & 0.608 \\
CPR w/o Const. & \textbf{0.800} & 0.273 \\
\midrule
\textbf{CPR (Full)} & 0.797 & \textbf{0.613} \\
\bottomrule
\end{tabular}
\end{sc}
\end{small}
\end{center}
\end{table}

\subsection{Safety Analysis}
Clinical AI deployment is often hindered by domain shifts. To evaluate the "safety" of our method, we conducted a zero-shot evaluation on the \textbf{Chapman-Shaoxing} dataset \cite{zheng2020chapman}.

\begin{table}[t]
\caption{\textbf{Zero-Shot Cross-Domain Safety.} CPR mitigates catastrophic failure under distribution shifts.}
\label{tab:cross_domain}
\begin{center}
\begin{small}
\begin{sc}
\begin{tabular}{lccc}
\toprule
Method & Clean F1 & SAP F1 & Drop \\
\midrule
Baseline & 0.546 $\pm$ 0.006 & 0.392 $\pm$ 0.020 & -28.3\% \\
\textbf{CPR} & 0.541 $\pm$ 0.003 & \textbf{0.437 $\pm$ 0.008} & \textbf{-19.3\%} \\
\bottomrule
\end{tabular}
\end{sc}
\end{small}
\end{center}
\end{table}

\begin{figure}[b]
\vskip 0.1in
\begin{center}
\includegraphics[width=\columnwidth]{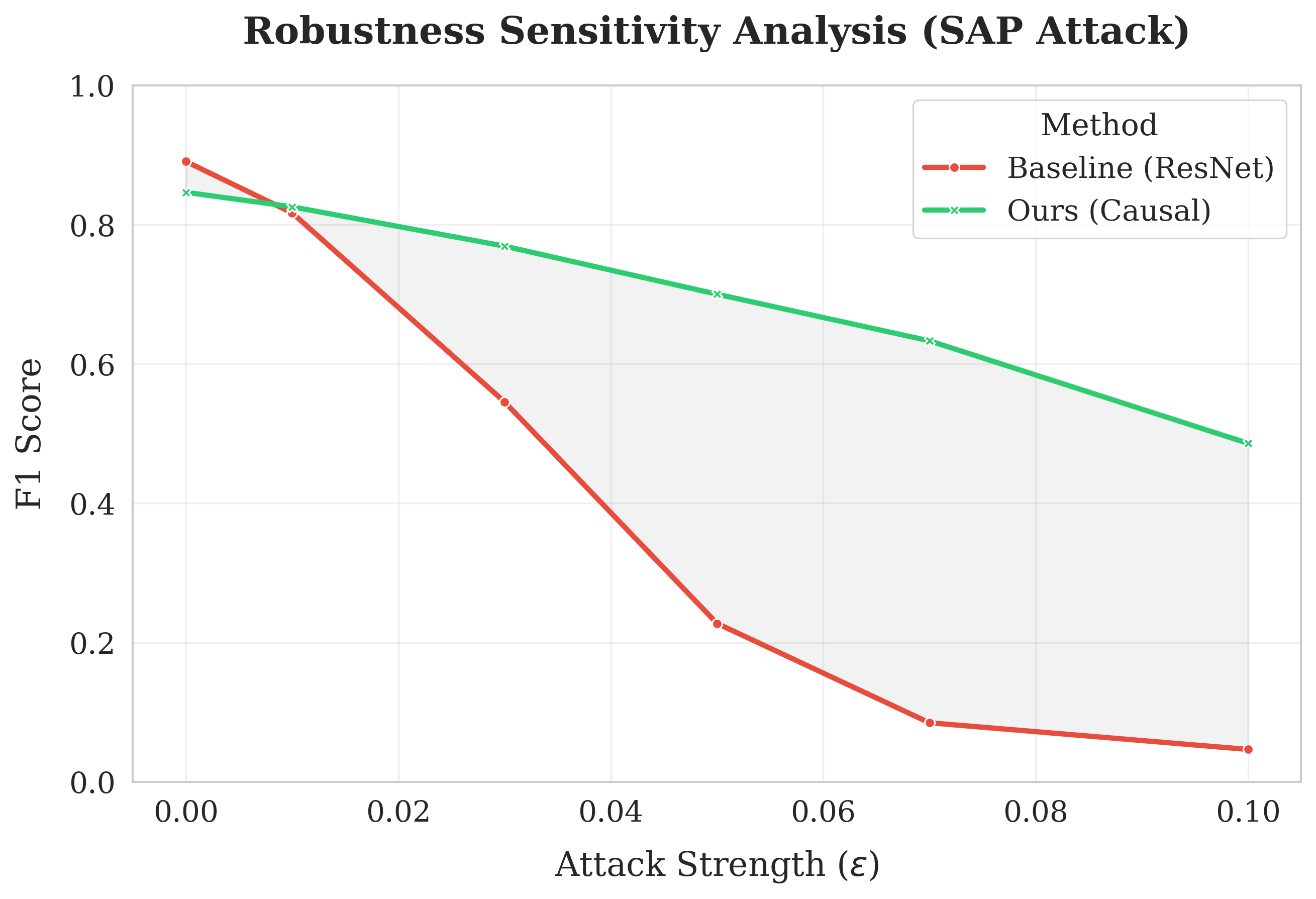}
\caption{\textbf{Robustness Sensitivity.} CPR (Green) maintains high F1 significantly longer than Baseline (Red) as $\epsilon$ increases.}
\label{fig:sensitivity}
\end{center}
\vskip -0.1in
\end{figure}

Table \ref{tab:cross_domain} shows zero-shot performance. CPR maintains a higher safety margin (F1 0.437) compared to Baseline (0.392) under attack. We also analyze sensitivity to perturbation strength in Figure \ref{fig:sensitivity}.

\section{Conclusion}
We presented CPR, a framework utilizing physiological priors for robust ECG analysis. By enforcing structural invariance, CPR provides a theoretically grounded and computationally efficient defense, matching certified methods in robustness while enabling real-time application.

\bibliography{example_paper}
\bibliographystyle{icml2025}

\newpage
\appendix
\onecolumn
\section{Theoretical Analysis and Proofs}
\label{app:proofs}

\subsection{Proof of Structural Invariance}
\textbf{Revised Assumption.} To rigorously establish structural invariance, we assume the training objective includes a regularization term on the encoder parameters $\theta_c$ (e.g., weight decay), promoting a minimum-norm solution. The total objective is:
\begin{equation}
    \mathcal{J}(\theta_c) = \mathbb{E}_{x} \left[ || (x - G(E_c(x; \theta_c))) \odot M ||^2 \right] + \lambda ||\theta_c||^2
\end{equation}

\begin{proof}
Let the encoder $E_c$ be parameterized by $\theta_c$. Ideally, we can decompose the encoder's dependency on the input $x$ into two subspaces defined by the mask $M$: the causal subspace $\mathcal{X}_M = \{x \odot M\}$ and the background subspace $\mathcal{X}_{\bar{M}} = \{x \odot (\mathbf{1}-M)\}$.

Since the reconstruction loss term $\mathcal{L}_{recon} = || (x - \hat{x}) \odot M ||^2$ only penalizes errors within the mask support, the gradient of the data fidelity term with respect to any input features in $\mathcal{X}_{\bar{M}}$ is strictly zero. 

Formally, let us consider a linear approximation of the encoder (or a single layer) $z = Wx + b$. We can decompose $W$ into $W_M$ (weights connected to causal features) and $W_{\bar{M}}$ (weights connected to background).
The gradient of the reconstruction loss $\mathcal{L}_{rec}$ with respect to $W_{\bar{M}}$ is:
\begin{equation}
    \frac{\partial \mathcal{L}_{rec}}{\partial W_{\bar{M}}} = \frac{\partial \mathcal{L}_{rec}}{\partial \hat{x}} \frac{\partial \hat{x}}{\partial z} \frac{\partial z}{\partial W_{\bar{M}}}
\end{equation}
Critically, because the loss is masked, $\frac{\partial \mathcal{L}_{rec}}{\partial \hat{x}_j} = 0$ for all $j$ where $M_j=0$. However, for the encoder to be truly invariant, we require $W_{\bar{M}}$ to converge to $\mathbf{0}$.

Without regularization, $W_{\bar{M}}$ would remain at its initialization values (random noise), and invariance would \textit{not} hold. However, under the assumption of the regularization term $\lambda ||\theta_c||^2$:
\begin{equation}
    \frac{\partial \mathcal{J}}{\partial W_{\bar{M}}} = \underbrace{\frac{\partial \mathcal{L}_{rec}}{\partial W_{\bar{M}}}}_{=0} + 2\lambda W_{\bar{M}} = 2\lambda W_{\bar{M}}
\end{equation}
Setting the gradient to zero for optimality ($\frac{\partial \mathcal{J}}{\partial W_{\bar{M}}} = 0$) implies:
\begin{equation}
    2\lambda W_{\bar{M}}^* = 0 \implies W_{\bar{M}}^* = \mathbf{0}
\end{equation}
Thus, at the global minimum under regularization, the weights connecting to the background region are driven to zero. Consequently:
\begin{equation}
    E_c^*(x + \delta) = W_M^* (x+\delta)_M + \underbrace{W_{\bar{M}}^*}_{=\mathbf{0}} (x+\delta)_{\bar{M}} = W_M^* x_M = E_c^*(x)
\end{equation}
This proves that the optimal encoder ignores any perturbation $\delta$ where $\delta \odot M = \mathbf{0}$.
\end{proof}

\end{document}